\documentclass[twoside]{article}

\usepackage[accepted]{aistats2019}
\usepackage{times}  
\usepackage{helvet}  
\usepackage{courier}  
\usepackage{url}  
\usepackage{graphicx}  
\usepackage{subcaption}
\usepackage{amsmath}
\usepackage{amssymb}
\usepackage{algorithm}
\usepackage[noend]{algpseudocode}
\usepackage[hidelinks]{hyperref}       
\usepackage{url}            
\usepackage{booktabs}       
\usepackage{amsfonts}       
\usepackage{nicefrac}       
\usepackage{microtype}      
\usepackage{xcolor}
\usepackage{multirow}
\usepackage{paralist}
\usepackage{dsfont}
\usepackage[export]{adjustbox}

\usepackage[style=numeric-comp,
            sorting=none,
            backend=bibtex,
            maxnames=2,
            maxbibnames=99]{biblatex}
\bibliography{refs}

\renewcommand{\vec}[1]{\ensuremath{\boldsymbol{#1}}}
\renewcommand{\v}[1]{\vec{#1}}

\newcommand{\x}{\ensuremath{\v{x}}}
\newcommand{\z}{\ensuremath{\v{z}}}

\renewcommand{\o}{\ensuremath{\omega}}
\newcommand{\p}{\ensuremath{\rho}}
\newcommand{\q}{\ensuremath{\theta}}
\newcommand{\f}{\ensuremath{\phi}}
\newcommand{\s}{\ensuremath{\v{\psi}}}

\renewcommand{\l}{\ensuremath{\v{\lambda}}}

\renewcommand{\b}{\ensuremath{\beta}}

\renewcommand{\L}{\ensuremath{\mathcal{L}}}
\newcommand{\E}{\ensuremath{\mathbb{E}}}

\newcommand{\N}{\ensuremath{\mathcal{N}}}

\begin{document}

%

%

\twocolumn[

\aistatstitle{Structured Neural Topic Models for Reviews}

\vspace{-1em}
\aistatsauthor{ Babak Esmaeili\\Northeastern University\\
  \texttt{esmaeili.b@husky.neu.edu} \And Hongyi Huang\\Viewpoint School\\
  \texttt{f.huang19@viewpoint.org} \AND   Byron C. Wallace\\Northeastern University\\
  \texttt{b.wallace@northeastern.edu} \And Jan-Willem van de Meent\\Northeastern University\\
  \texttt{j.vandemeent@northeastern.edu} }

\aistatsaddress{} 
]

\begin{abstract}
\vspace{-0.5em}
We present Variational Aspect-based Latent Topic Allocation (VALTA), a family of autoencoding topic models that learn aspect-based representations of reviews. VALTA defines a user-item encoder that maps bag-of-words vectors for combined reviews associated with each paired user and item onto structured embeddings, which in turn define per-aspect topic weights. We model individual reviews in a structured manner by inferring an aspect assignment for each sentence in a given review, where the per-aspect topic weights obtained by the user-item encoder serve to define a mixture over topics, conditioned on the aspect. The result is an autoencoding neural topic model for reviews, which can be trained in a fully unsupervised manner to learn topics that are structured into aspects. Experimental evaluation on large number of datasets demonstrates that aspects are interpretable, yield higher coherence scores than non-structured autoencoding topic model variants, and can be utilized to perform aspect-based comparison and genre discovery.
\end{abstract}

\section{Introduction}
\vspace{-0.5em}
In recent years the field of natural language processing (NLP) has decisively shifted away from bag-of-words representations towards neural models. These models represent text using embeddings that are learned from data in an end-to end manner. A potential drawback to such embeddings is that learned representations tend to be \emph{entangled}, in the sense that an embedding is a monolithic vector that encodes some unknown set of characteristics of the input data. When one is interested in training a model solely for a particular task, entanglement is not necessarily a problem, so long as the trained model achieves sufficiently robust predictive performance. However, there are cases where it is desirable to learn a representation that factors into distinct, complementary sets of features, i.e., is \emph{disentangled}. 

One reason we may want a disentangled representation is interpretability. Separating representations into distinct factors that correspond to identifiable subsets of features, such as the topic and political leaning of an opinion piece, allows one to more easily reason about which features informed a prediction. A second reason to induce disentangled representations is data efficiency. Suppose that were to train a model on images that contain $K$ categories of shapes which assume $L$ categories of colors. If a model can separate shape from color, then it should generalize to shape and color combinations not observed in the training data. This means that we can hope to train such a model on $O(K + L)$ examples, rather than a dataset in which all combinations of features are present, which would require $O(KL)$ examples. Learning disentangled representations thus provides a strategy for factorizing a problem in a high-dimensional feature space into problems in lower-dimensional feature spaces.

In computer vision, there has been considerable effort to develop methods for inducing disentangled representations in semi- and un-supervised settings \cite{kingma2013auto-encoding,higgins2016beta,siddharth2017learning,esmaeili2018structured,zhao2017infovae,gao2018auto,achille2018information,kim2018disentangling,chen2018isolating}. 
Many of these approaches define deep generative models such as variational autoencoders (VAE) \cite{kingma2013auto,rezende2014stochastic}, or Generative Adversarial Networks (GANs) \cite{goodfellow2014generative,chen2016infogan}. In NLP, work on learning disentangled representations has been more limited \cite{ruder2016hierarchical,he-2017,zhang2017aspect,jain-EMNLP-18}. A large body of pre-neural work exists on aspect-based topic models that derive from Latent Dirichlet Allocation (LDA) \cite{blei2003latent}. This includes approaches for sentiment analysis \cite{brody2010unsupervised,sauper2010incorporating,sauper2011content,mukherjee2012aspect,sauper2013automatic,kim2013hierarchical}, and models in the factorial LDA family \cite{paul2010two,paul2012factorial,wallace2014large}. 

There has been relatively little work in NLP on learning disentangled representations with neural architectures. 
One reason for this is that work on deep generative models for text is not as well-established as work for images. 
Early approaches in this space, such as the Neural Variational Document Model (NVDM) \cite{miao2016neural} and autoencoding LDA \cite{srivastava2017autoencoding} developed neural topic models in which the generative model is either an LDA-style mixture, or a SAGE-style \cite{SAGE} log-linear combination over topics. 
More recently there have been some efforts to develop deep generative models with interpretable aspects, chiefly the work by \textcite{hu2017controlled}, which combines a recurrent VAE architecture with a set of aspect discriminators to induce a structured representation.



In this paper we explore the effectiveness of neural topic models for learning disentangled representations. We treat review datasets as a particular case study, where we consider the task of learning structured representations for both the reviewer and the reviewed item. Reviews comprise several variables of interest, such as the aspect of the item being discussed, user sentiment regarding each aspect, and characteristics of the item for each aspect (i.e., sub-aspects). More concretely, in any review corpus, items will very likely share certain \emph{aspects}, each affecting the rating separately. For example, in the case of restaurant reviews, all establishments will serve food and have a location. Similarly, every beer will have an aroma and appearance. More generally, each aspect may contain nested \emph{sub-aspects}: A restaurant can serve Italian, Chinese, or fast food; and a beer can be dark or light in appearance. 

In this paper we develop autoencoding models that induce representations of review texts that capture this structure. Such representations can perform aspect-based item comparisons, and also provide one sort of interpretability. 
To realize these goals, VALTA combines topic modelling and recommender systems into a structured VAE framework. We model reviews in a structured manner by associating an aspect with each sentence in a review, and use aspect-specific topics to define a log-linear likelihood, similar to the one used in SAGE \cite{SAGE}, the NVDM \cite{miao2016neural}, and ProdLDA \cite{srivastava2017autoencoding}. Topic and aspect weights are predicted based on a user and item embedding. The result is a highly structured model, in which both aspects and sub-aspects are interpretable, and topics have a high predictive power in terms of perplexity and coherence scores. These learned representations can be used for downstream tasks such as genre discovery, representation quality, and aspect-retrieval. 

\begin{table}[t]
    \centering
    \begin{tabular}{ll}
        \hline
        Symbol      & Description   \\
        \hline
        $V$         & vocabulary size \\
        $A$         & number of aspects \\
        $K$         & number of sub-aspects \\
        $H$         & number of hidden units \\
        $u$         & user index \\ 
        $i$         & item index \\
        $\q$        & parameters of generative model \\
        $\f$        & parameters of inference model \\
        $\x_{i,u}$  & review written by user $u$ about item $i$ \\
        $\x_{i,u,s}$& sentence $s$ of $\x_{i,u}$ \\
        $\o_{i,u,s}$& aspect log probabilities of $\x_{i,u,s}$\\
        $\z_{i,u,s}$& aspect assignment of $\x_{i,u,s}$ \\
        $\p_{i}$    & hidden representation of item $i$ \\
        $\p_{u}$    & hidden representation of user $u$ \\
        $\p_{i,u}$  & hidden representation of $\x_{i,u}$ \\
        $\s_{i,u}$  & aspect-specific topic distributions of $\x_{i,u}$ \\
        $\l_{i}$    & aspect-importance of item $i$ \\
        $\l_{u}$    & aspect-preference of user $u$ \\
        $\b_{0}$    & global rating bias \\
        $\b_{i}$    & item rating bias \\
        $\b_{u}$    & user rating bias \\
        $r_{u,i}$   & true rating by user $u$ for item $i$ \\
        $\hat{r}_{u, i}$ & prediction of rating by user $u$ for item $i$ \\
        \hline
    \end{tabular}
    \caption{Summary of notation used throughout.} 
    \label{tab:notatation-table}
\end{table}

\section{Background and Preliminaries}

\begin{figure*}[!t]
    \centering
    \includegraphics{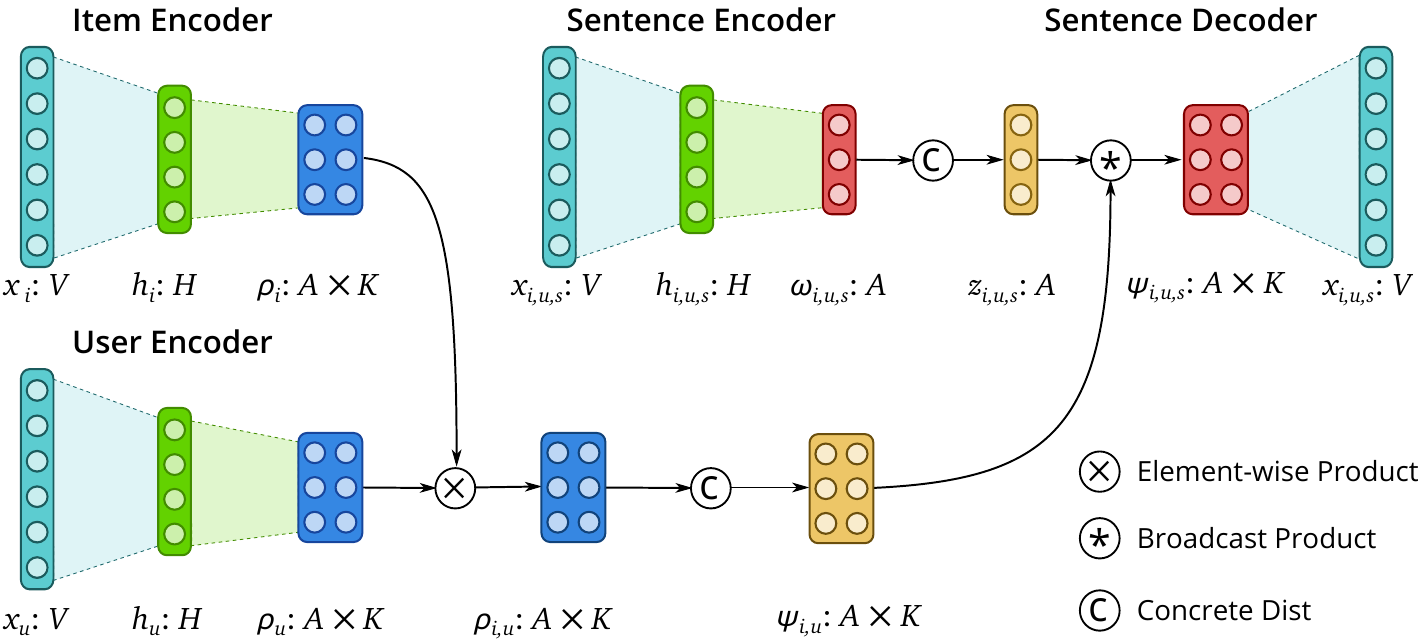}
    \caption{VALTA learns representations for reviews using a structured autoencoder that consists of a sentence-level 
    encoder $q_\phi(\z_{i,u,s} \mid \x_{i,u,s})$, which infers assignments to aspects, and a document-level user-item encoder $q_\phi(\s_{i,u} \mid \x_{i}, \x_{u})$, which infers per-aspect topic weights, and a sentence-level decoder $p_\theta(\x_{i,u,s} \mid \z_{i,u,s}, \s_{i,u})$.}
    \label{fig:VALTA}
\end{figure*}

Review datasets have been widely studied in the context of recommender systems \cite{bennett2007netflix}. Matrix factorization techniques \cite{koren2009matrix,mnih2008probabilistic,bao2014topicmf} are widely used to predict ratings by representing each user and item by a $K$ dimensional vector, which we sometimes refer to as an embedding. Since these approaches consider the ratings alone, they ignore the text of the review, which is a key source of information. \textcite{mcauley2013hidden} proposed combining topic models and matrix factorization techniques for learning ratings to learn topics and ratings simultaneously. Subsequent approaches aimed to exploit review text in addition to rating \cite{mcauley2013hidden, diao2014jointly, zheng2017joint, catherine2017transnets, cheng2018aspect}. These efforts have shown that topic models indeed can act as a good regularizer for rating prediction, particularly for users or items with few reviews \cite{mcauley2013hidden, cheng2018aspect}. In the last few years, both recommender systems and topic modelling approaches have shifted towards deep learning methods \cite{srivastava2017autoencoding, miao2016neural, diao2014jointly}, many of which also exploit text to predict ratings.

While neural recommender systems can achieve good predictive performance, it is unclear how they do so, because learned feature vectors are optimized only to code indiscriminately for (unknown) predictive combinations of attributes. Such entangled representations thus do not reveal any information about the structure of the data, which in turn hinders model interpretability and generalizability. By imbuing representations with probabilistic semantics, we can design the models to explicitly tease out structured embeddings, components of which may then be re-used. 

In prior work, deep generative models have been proposed to learn representations of text via variational autoencoders \cite{kingma2013auto, rezende2014stochastic}. VAEs jointly optimize a generative network and an inference network. The former, $p_{\q}(\x, \z)$, specifies a distribution over set of hidden variables $\z$ and observed variables $\x$. The latter is a conditional distribution $q_{\f}(\z | \x)$. Defining $q(x)$ as the empirical distribution, these two models are trained by optimizing the evidence lower bound (ELBO),
\begin{align}
    \label{eq:vae-elbo}
    \begin{split}
     \L(\q, \f)
    :&=
    \E_{q(\x)}
    \left[
    \E_{q_{\f}(\z | \x)}
    \left[
    \log
    \frac{p_{\q}(\x, \z)}{q_{\f}(\z \mid \x)}
    \right]
    \right].
    \end{split}
\end{align}
Variants of VAEs have been used to develop autoencoding topic models \cite{srivastava2017autoencoding, miao2016neural}. These models achieve predictive performance (in terms of the perplexity score) that is competitive with other bag-of-word models, but lack the explicit structure that we aim to capture here. More specifically: the prior in existing VAE-based approaches takes the form of a Gaussian with diagonal covariance matrix, where each dimension of the Gaussian corresponds to a topic. By contrast, we here aim to characterize groups of topics that correspond to specific aspects of interest. One means of realizing this would be to posit $A$ Gaussians, one per aspect, and each of these might comprise its own set of $K$ topics.

\section{Methodology}

To learn structured representations of reviews, we begin by identifying key axes of variation in review datasets. We define three variable categories: items, users, and review texts. We assume $A$ aspects of interest for all items. For example, in the case of beer reviews, these aspects correspond to properties such as appearance and aroma. We further decompose these aspects into $K$ topics. A topic within appearance might be, e.g., dark versus pale beer. Reflecting these structural assumptions, our model defines aspect-specific embeddings that in turn yield distributions over topics. Thus, a representation of a sweet, dark beer should place a relatively large mass on the dark topic of appearance, and a large mass on the sweet topic of appearance.

The relative importance of aspects may vary for both items and users. A restaurant, for example, may be located on the water or on a famous city street, in which case the location is likely to be its most salient aspect. Similarly, lagers are not typically renowned for their smell. Users will have their own weightings of aspect importance. A particular user may be concerned primarily with food quality over price, and might prefer Chinese food. Others, meanwhile, may emphasize location or ambience. This sort of structure is is similar to the aspect-aware topic model proposed in \cite{cheng2018aspect}.


The words contained in a review are a function of the aspects and topics, and their relative importance for particular user-item \emph{pairs}. To accurately learn topics and predict ratings, we now introduce variables that are defined at the review level. A naive approach would be to encode the review text $\x_{i,u}$ and then train the generative model to learn both topics and ratings for item $i$ and user $u$. Here we propose an approach that is directly motivated by the observation that the aspects and topics discussed in review $\x_{i,u}$ will depend on a combination of the aspect preferences of $u$, and the relative salience of the respective aspects for $i$. Therefore, rather than encoding $\x_{i,u}$ directly, we encode information about $i$ and $u$ separately, and then combine these representations to yield a joint embedding for $\x_{i,u}$ and predict the rating $r_{i,u}$. Table \ref{tab:notatation-table} presents the notation we use throughout this paper.

It is likely that most reviews will contain at least some words about all aspects (although the prevalence of individual aspects will vary across reviews). Thus it is intuitive to attempt to infer which parts of a review talk about which aspect. In our model we make the simplifying assumption that every sentence within a review discusses only a single aspect. One could alternatively assign aspects at the word- or paragraph-level. However, sentence-level assignment constitutes an intuitive compromise, and is also consistent with prior work \cite{mcauley2012learning,lu2011multi}. Note that while the aspect assignment varies between sentences within a review, we keep the topic proportions fixed for that particular review. 

Following prior work \cite{mcauley2013hidden, diao2014jointly, cheng2018aspect}, we assume the input representation for item $i$ is a bag-of-words vector encoding the words used across all reviews written about this item. Similarly, we define the input vector for user $u$ as a bag-of-words induced over all reviews that they have written. This representation has been shown to perform well in terms of capturing characteristics of items and users \cite{mcauley2013hidden,catherine2017transnets,zheng2017joint}, but it does not take into account the relative importance of different aspects with respect to both $i$ and $u$. Nor have such models explicitly accounted for the intuitive observation that different parts of reviews (probably) discuss different aspects, which we achieve via sentence-wise aspect assignments $\{\z_{i,u,s}\}$ based on encoded sentences $\{\x_{i,u,s}\}$ (one per each sentence in the review written by $u$ for $i$). 

We provide a schematic of our model in Figure \ref{fig:VALTA}. The inference and generative models are defined to codify the structure discussed above. Specifically, given the topic distribution $\s_{i,u}$ for item $i$ and user $u$, sentence aspect assignments $\{\z_{i,u,s}\}$, and the review $\x_{i,u}$, we define the inference model
\begin{align}
    \label{eq:q-distribution}
    \begin{split}
    &q_{\f}(\s_{i,u}, \z_{i,u} 
    \mid 
    \x_{i}, \x_{u}, \x_{i,u}) 
    \\
    &~~=
    q_{\f}(\s_{i,u}
    \mid
    \x_{i},\x_{u})
    \prod_{s}
    q_{\f}(\z_{i,u,s}|\x_{i,u,s}).
    \end{split}
\end{align}
An obvious choice for the likelihood model is to define a decoder for review $\x_{i,u}$. However, this will entangle the different aspects discussed in said review. To ensure that the generative model associates different dimensions with specific aspects, we define our generative network at a sentence level
\begin{align}
\label{eq:p-distribution}
\begin{split}
&p_{\q}(\x_{i,u}, \z_{i,u}, \s_{i,u}) 
\\
&=
\prod_{s}
p_{\q}(\x_{i,u,s}| \z_{i,u,s}, \s_{i,u})
p(\z_{i,u,s})
\prod_{a}
p(\s_{i,u,a}) .
\end{split}
\end{align}
We note that the $K$-dimensional topic distribution $\s_{i,u}$ is fixed at the review level. This reflects the assumption that given the item and user, the specific topics of interest will not change, as the opinion of the user and the characteristics of the item are fixed. The only axis of variation is the user's decision regarding which aspect to write about in any given sentence. However, we must ensure that the generative model focuses only on the assigned aspect, rather than topics of all aspects. We enforce this by multiplying the columns of $\s_{i,u}$ with the (nearly) one-hot topic assignment vector $\z_{i,u,s}$: $\s^{s}_{i,u}=\z_{i,u,s}\s_{i,u}$. Because $\z_{i,u,s}$ resembles a one-hot vector, this effectively masks the topic distributions pertaining to other (unassigned) topics. Thus, only the topic distribution corresponding to a single (selected) aspect is responsible for reconstructing $\x_{i,u,s}$. 

We use the generative model for text introduced in prior work \cite{srivastava2017autoencoding}. This model induces $\log$ probabilities for each word via feeding the $\s^{s}_{i,u}$ through a single layer neural network followed by applying a log softmax
\begin{align*}
    \log
    p_{\q}(\x_{i,u,s}
    |
    \s^{s}_{i,u})
    &=
    \x_{i,u,s}
    \log
    \left(
    \frac{\text{exp}(\hat{w}_{v})}
    {\sum_{v=1}^{V}\text{exp}(\hat{w}_{v})}
    \right).
\end{align*}
Where the log word probabilities $\hat{w} = g_{\q}(\s^{s}_{i,u})$ are computed using a single-layer perceptron $g_{\q}$ with weights $\q$.

\begin{table*}[t!]
    \centering
    \begin{tabular}{llllll}
        \textbf{Dataset}    & \textbf{Aspects}                                      & \textbf{\#users}  & \textbf{\#items}  & \textbf{\#reviews}      &   \textbf{\#sentences}            \\
        \hline 
        Beer (Beeradvocate) & Aroma, Taste, Mouthfeel, Look                         & 4,923             & 2,017              & 127,346               &   1,515,517                       \\
        Restaurant (Yelp)   & Price, Ambiance, Food, Service                        & 13,847            & 6,588              & 140,139               &   1,416,317                       \\
        Clothing (Amazon)   & Formality, Appearance, Type                           & 12,203            & 73,903             & 80,285                &   447,920                         \\
        Movie (Amazon)      & Genre, Awards, Screen Play                            & 7,590             & 2,288              & 100,489               &   1,446,690
    \end{tabular}
    \caption{Review dataset statistics.}
    \label{tab:datastes}
\end{table*}

\subsection{Concrete Distribution}
An important factor in our model is the choice of $p(\z)$ and $p(\s)$. The appropriate choice for $p(\z)$ and $p(\s)$ are discrete and Dirichlet distributions respectively, as they represent aspect \emph{assignment} and topic \emph{proportions}. This is problematic in practice because discrete variables are not amenable to the reparameterization trick, thus precluding use of estimation via standard backpropagation algorithms. In the case of Dirichlet distributions, several methods have been proposed to allow for sampling via reparameterization \cite{ruiz2016generalized, figurnov2018implicit}. However, in practice these methods dramatically increase training in our implementation because the base system, PyTorch, does not provide GPU implementations for these distributions at the time of writing. In this work we choose to model both variables $\z_{i,u,s}$ and $\s_{i,u,a}$ using the Concrete distribution, a relaxation of discrete distributions implemented via a Gumbel softmax \cite{maddison2016concrete, jang2016categorical}. The Gumbel distribution can be sampled in a reparameterized way by drawing $u \sim \text{Uniform}(0, 1)$ and then computing $g = − \log(− \log(u))$. If $\z$ has aspect log-probabilities $\o_1, \o_2, \ldots, \o_A$, then we can sample from a continuous approximation of the discrete distribution by sampling a set of $g_{a} \sim \text{Gumbel}(0, 1)$ i.i.d.~and applying the transformation
\begin{equation*}
    \z_a
    =
    \frac{{\exp}((\o_a + g_a)/\tau)}
    {\sum_{a}\exp((\o_a + g_a)/\tau)}
    .
\end{equation*}
where $\tau$ is a temperature parameter controlling relaxation. The sample $\z$ is a continuous approximation of the desired one-hot representation. The role of $\tau$ is critical in our model, as it dictates the \emph{peakiness} of the samples. In the case of $\z_{i,u,s}$, we keep the temperature low to enforce the assumption that each sentence is only talking about a certain aspect. However, we do not wish for the topic proportions to be close to a one-hot vector, as this would restrict items to contain a single topic within each aspect. To encourage $q_{\f}(\s)$ to mimic a Dirichlet distribution, we set the temperature to higher values, thereby encouraging all $K$ dimensions within each aspect to contribute. In our experiments, we have observed that sampling $\s_{i,u}$ with a low temperature results in few dimensions within each aspect learning something meaningful about the review. 

\subsection{Rating Prediction}

A good representation of a review should not contain only informative topics, but should also assist in accurately predicting the rating linked to the review. In this subsection, we extend VALTA to predict ratings in combination with learning aspects and topics. We take several factors into account when predicting the rating. As discussed above, we assume that users have different aspect preferences and that items exhibit different aspect-importance. To extract aspect-importance vectors $\l_{i}$ and $\l_{u}$ for item $i$ and user $u$, respectively, we use the weights of the sentence encoder $f^{\textit{aspect}}_{\f}(h)$ that is responsible for predicting the aspect
\begin{equation*}
    \l_{i} = f^{\textit{aspect}}_{\f}(h_i), \qquad \l_{u} = f^{\textit{aspect}}_{\f}(h_u).
\end{equation*}
As $f^{\textit{aspect}}_{\f}(\cdot)$ is trained at the sentence level (and so compelled to extract words associated with aspects), we can re-use its weights to extract an aspect-importance vector from a collection of reviews. We then average these two embeddings to obtain aspect-importance for a particular pair 
\begin{equation*}
     \l_{i,u} = \frac{1}{2}(\l_{i} + \l_{u})
\end{equation*}
One could consider learning different embeddings for items and users that are different from the one in topic models. However, as discussed in \cite{mcauley2013hidden}, \emph{coupling} the embeddings for items and users with their topic models representation helps to learn topics that explain the diversities in ratings. Thus, in our model we re-use the input to the concrete distribution $\p_{i,u}$ to predict the rating associated with each aspect as
\begin{equation*}
    \hat{r}_{i,u,a} = \sum_{k=1}^{K} \p_{i,u,a,k}.
\end{equation*}
Based on this structure we predict the overall ratings as
\begin{equation}
\label{eq:rating-prediction}
    \hat{r}_{i, u} = \b_{0} + \b_{i} + \b_{u} + \frac{1}{A}\sum_{a=1}^{A} \underbrace{\l_{i,u,a}}_{\textit{aspect importance}} 
    \!\!\!\!\!\!\!\!
    \overbrace{\hat{r}_{i,u,a}}^{\textit{aspect rating}}
\end{equation}
where $\b_{0}$ is the global rating bias, and $\b_{i}$ and $\b_{u}$ are item $i$ and user $u$ bias respectively. This approach is similar to the family of aspect-aware latent factor models (ALFM) proposed in \cite{cheng2018aspect}. Following prior work, we use the  mean squared error (MSE) loss for the recommender model component. This may also be interpreted in a probabilistic way where we model $\hat{r}_{u, i}$ as a Gaussian distribution:
$p(\hat{r}_{u, i}; r_{u,i}) := \N(\hat{r}_{i, u}; r_{i, u}, 1)$. 

\subsection{VALTA}

In this subsection, we put everything together to define a unified objective for VALTA. For clarity, we decompose our objective into the four terms, which together define a lower bound on the log marginal likelihood, analogous to the VAE objective defined in equation \ref{eq:vae-elbo}
\begin{equation}
\label{eq:VALTA}
    \L^{\textit{VALTA}} (\q, \f) = 
    \L^{\x}_{gen}
    +
    \L^{r}_{mse}
    +
    \L^{\s}_{\textit{KL}}
    +
    \L^{\z}_{\textit{KL}}
    .
\end{equation}
The first term is the expected log likelihood of the review
\begin{equation*}
    \L^{\x}_{\textit{gen}}(\q, \f)
    =
    \E
    \left[ 
    \log 
    \prod_{s}
    p_{\q}(\x_{i,u,s} | \s_{i,u}, \z_{i,u,s})
    \right].
\end{equation*}
Note that this expectation is are defined w.r.t to the inference model $q_{\f}(\cdot)$, which we omit for simplicity. 

The second term is the likelihood of the rating $r_{i,u}$
\begin{equation*}
    \L^{r}_{\textit{mse}}(\q)
    =
    \log p_{\q}(r_{i,u} | \x_{i}, \x_{u})
    .
\end{equation*}
Finally, as with a normal VAE we incorporate two regularization terms in the form of KL divergences between the encoder distribution and the prior
\begin{align*}
    \L^{\z}_{\textit{KL}}(\q, \f)
    &= 
    -\E
    \left[ 
    \log 
    \prod_{s}
    \frac{q_{\f}(\z_{i,u,s} | \x_{i,u,s})}
    {p_{\q}(\z_{i,u,s})}
    \right],
    \\
    \L^{\s}_{\textit{KL}}(\q, \f)
    &= 
    -\E
    \left[ 
    \log 
    \frac{q_{\f}(\s_{i,u}|\x_{i}, \x_{u})}
    {p_{\q}(\s_{i,u})}
     \right]
     .
\end{align*}
where the terms are responsible for reconstructing the review text, predicting the rating, matching the aspect distribution in the encoder to prior, and matching the topic distributions in the encoder to the prior respectively. 

\section{Related Work}

A comprehensive literature review of recommender systems is beyond the scope of this work. Here we discuss models that exploit both rating and reviews to jointly learn topics and predict ratings. We divide these models to three classes: 1) probabilistic topic models; 2) deep learning-based approaches; 3) VAEs. VALTA belongs to the last category.

In the first class, the most closely related approach to our work is the aspect-aware topic model (ATM) \cite{cheng2018aspect}, which considers a similar decomposition of reviews to aspects and sub-aspects. In the same paper, the authors also propose an aspect-aware latent factor model (ALFM) which exploits the parameters learned from the ATM to predict ratings. While VALTA shares the idea of further decomposition of aspects with ATM, it is trained to learn topics and predict ratings jointly rather than sequentially. 

Another related model is factorial-LDA  \cite{NIPS2012_4784} which learns a facorized topic structure. Their approach to learn structured topics is different to VALTA in that factorial-LDA learns topics as \emph{tuples} while VALTA learns topics as hierarchies. Other approaches similar to ours are \cite{mcauley2013hidden,diao2014jointly,zhao2015improving,zhang2014explicit}. However, they are all purely probabilistic models and furthermore they do not consider hierarchical topics. 

\begin{table}[!t]
    \centering
    \begin{tabular}{c|cc|cc}
                &  \multicolumn{2}{|c}{\textbf{CitySearch}}     & \multicolumn{2}{|c}{\textbf{BeerAdvocate}}    \\
                &  ACC          & F1-Score                       & ACC                      & F1-Score              \\
    \hline 
    \hline
    LDA         & 0.477          & 0.597                          & 0.447                   & 0.468                  \\
    Local-LDA   & 0.803          & 0.861                          & 0.758                   & 0.761                  \\
    SVM         & 0.830          & 0.887                          & 0.647                   & 0.604                  \\
    \hline
    VALTA       & \textbf{0.885} & \textbf{0.931}                & \textbf{0.769}          & \textbf{0.794}         \\
    \end{tabular}
    \caption{Multi-aspect sentence labelling results.}
    \label{tab:masl-results}
    \vspace{-1em}
\end{table}

Two recent, closely related deep learning-based methods are \cite{catherine2017transnets, zheng2017joint}. Both exploit the review texts for a pairs $(i, u)$ to predict ratings. While these models perform well in terms of predicted ratings, they are not designed to learn topics. VAEs have also been used for collaborative filtering \cite{li2017collaborative, karamanolakis2018item}. However, as pointed out by \cite{liang2018variational}, these approaches tend to under-fit the data. In the vision domain, the idea of capitalizing on more complex priors in VAEs has become a popular idea and has been strongly associated with disentanglement \cite{kingma2014semi,narayanaswamy2017learning,esmaeili2018structured}. However, this idea has not been emphasized as much in natural language processing.

To our knowledge, VALTA is the only VAE-based approach that considers hierarchical topics. Furthermore, our encoder architecture is unique in that it couples a sentence-level decoder with an item and user encoder. We also note that aspect classification has also been separately studied in the context of semantic analysis \cite{poria2016aspect,mcauley2012learning,lu2011multi,schouten2016survey}.

\vspace{-0.5em}
\section{Experiments}
\vspace{-0.5em}

To assess the structured representation learned by VALTA, we evaluate a number of tasks and datasets. The experiments are designed to evaluate the quality of aspects and topics, rating-prediction, and structure of the representations. We implement all VAE-based models in Probabilistic Torch \cite{siddharth2017learning}, a library for deep probabilistic models. In all experiments, we set the temperature for $\z_{i,u,s}$ and $\s_{i,u,a}$ to 0.66 and 5.0 respectively. The number of hidden units $H$ for all models is 256, followed by a $\tanh$ function. We ran all experiments for 200 epochs with batch-size 100 (at the review level). The optimizer we used was ADAM with default hyperparameters. We use $(A=5, K=2)$ for the beer review and $(A=10,K=3)$ for all other datasets.

The main findings in our experiments are as follows.
\vspace{-0.5em}
\begin{enumerate}
    \item VALTA can disentangle aspects of a review in a fully unsupervised manner. We demonstrate this on the CitySearch and BeerAdvocate datasets, which have been annotated with aspect-specific ratings \cite{mcauley2012learning, ganu2009beyond}. 
    \item VALTA learns word distributions for every aspect and topic that have a higher coherence score \cite{lau2014machine} than baseline methods at both the sentence and review level. This indicates interpretability. 
    \item VALTA learns a representation that can be used to make aspect-based comparisons of items and users.
    \item In all but one dataset, VALTA produces the most accurate rating predictions of all models considered.
\end{enumerate}

\begin{table}[!b]
    \centering
    \begin{tabular}{c|cc}
                    &   \multicolumn{2}{|c}{\textbf{BeerAdvocate}}     \\
                    & Sentence          & Review                                          \\
         \hline 
         \hline
        LDA         & \_                & 0.5756                                            \\
        Local-LDA   & 1.5196            & \_                                                    \\
        NVDM        & \_                & 0.2338                                            \\
        ProdLDA     & \_                & 0.2208                                            \\
        \hline 
        VALTA       & \textbf{1.817}    & \textbf{0.655}                                    \\
    \end{tabular}
    \caption{Average Topic Coherence (NPMI). We compare either the sentence-level or review-level coherence score, depending on the input-level employed in the baseline.}
    \label{tab:NPMI-table}
\end{table}

\begin{table*}[t]
\centering
\begin{tabular}{c|c||c|c||c|c||c|c}
    \multicolumn{2}{c||}{\textbf{Appearance}} & \multicolumn{2}{c||}{\textbf{Aroma-Taste} } & \multicolumn{2}{c||}{\textbf{Palate}} & \multicolumn{2}{c}{\textbf{Semantic}} \\
     \hline
     \hline
     golden & black & roasted   & citrus    & mouthfeel     & mouthfeel     & lagers    & try        \\
     yellow & tans  & coffee    & grapefruit& bodied        & watery        & heineken  & hype       \\
     white  & glass & vanilla   & pine      & smooth        & rjt           & macro     & recommend  \\
     orange & pour  & chocolate & hops      & carbonation   & bodied        & import    & overall    \\
     hazy   & head  & bourbon   & lemon     & medium        & refreshing    & euro      & founders   \\
     color  & pitch & oak       & floral    & drinkability  & carbonation   & lager     & favorite   \\
     gold   & lacing& malts     & clove     & drinkable     & crisp         & bmc       & stouts     \\
     copper & color & sweet     & malt      & alcohol       & dry           & worse     & stout      \\
     straw  & brown & aroma     & aroma     & finish        & finish        & bad       & ipa        \\
     amber  & ginger& malt      & grass     & mouth         & thirst        & skunky    & cheers     \\
\end{tabular}

\vspace*{1em}
\begin{tabular}{c|c|c||c|c||c||c|c}
    \multicolumn{3}{c||}{\textbf{Food}} & \multicolumn{2}{c||}{\textbf{Service}} & \multicolumn{1}{c||}{\textbf{Ambiance}} & \multicolumn{2}{c}{\textbf{Payment}} \\
     \hline
     \hline
     rice   & pepperoni & bagel         & service   & friendly      & walls     & card      & minutes \\
     chicken& provolone & eggs          & friendly  & service       & love      & debit     & card       \\
     sauce  & mozzarella& hash          & staff     & staff         & restaurant& stamp     & seated\\
     shrimp & bagel     & scrambled     & attentive & baristas      & located   & minutes   & table\\
     pork   & onions    & brown         & food      & helpful       & wall      & receipt   & debit    \\
     fried  & mushrooms & ham           & helpful   & employees     & decor     & cash      & waited   \\
     beef   & arugula   & lox           & server    & customer      & hidden    & register  & asked  \\
     noodles& knots     & benedict      & atmosphere& barista       & ceiling   & cards     & wait\\
     spicy  & artichoke & poached       & great     & clean         & gem       & order     & gratuity \\
     salmon & cheese    & capers        & knowledgeable& rude       & lighting  & cashier   & told     \\
\end{tabular}
\caption{Top 10 words learned by VALTA: beer (top) and restaurants (bottom).}
\label{tab:top-words}
\end{table*}

\vspace{-1.0em}
\subsection{Datasets and Preprocessing}

A summary of the data we use in this paper is shown in Table \ref{tab:datastes}. We focus on datasets that exhibit clear structure. We chose the \emph{BeerAdvocate} dataset\footnote{\url{http://snap.stanford.edu/data/}} and restaurant data from \emph{Yelp Dataset Challenge}\footnote{\url{https://www.yelp.com/dataset_challenge/}} from as they both contain explicit aspects. For the other two datasets, we selected Clothing and Movie reviews from Amazon \cite{mcauley2015image, he2016ups}. 

We preprocessed the datasets as follows. we used the Spacy library to remove all stop words, and we removed all words that occurred fewer than five times. We also used Spacy for sentence segmentation. The resultant vocabulary size for these datasets varies from 20000 to 30000. We also filtered reviews such that we include only items and users for which we have at least five reviews.

\begin{figure*}[!t]
    \centering
    \begin{minipage}{0.15\textwidth}
    \begin{tabular}{c|c}
        Model       & ACC  \\
        \hline
        \hline
         LDA        & 0.13  \\
         NVDM       & 0.48  \\
         ProdLDA    & 0.45  \\
         \hline
         VALTA      & \textbf{0.53}
    \end{tabular}
    \end{minipage}
    \hspace*{4ex}
    \begin{minipage}{0.80\textwidth}
    \includegraphics[width=\textwidth]{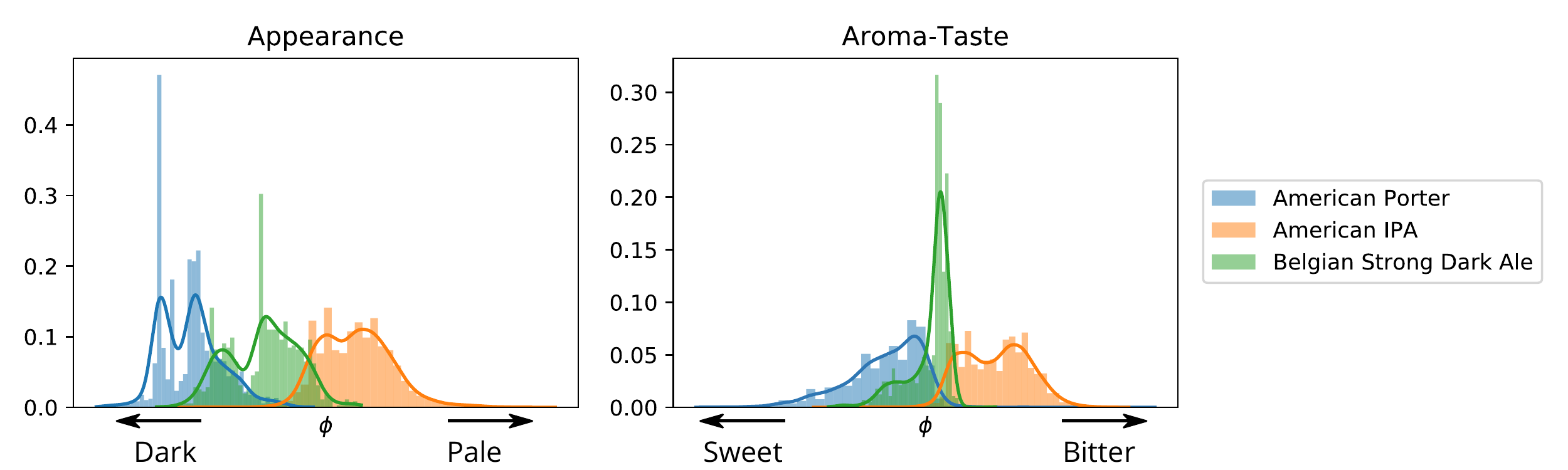}
    \end{minipage}
    \caption{Left: category classification table on beer data. Right: normalized density of latent values $\s_{i,u,a}$ of the encoded reviews of three different style of beer. }
    \label{fig:aspect-clustering}
\end{figure*}

\subsection{Baselines}

We compare our model to a diverse set of baseline models, including probabilistic, VAE-based, and aspect-based models. We also include a simple version of our own model that we call variational review LDA (VRLDA) which follows the implementation of VALTA save for the aspects. In other words, the representation used VRLDA is only a flat $K$ dimensional vector. The full list of baselines are: LDA \cite{blei2003latent}, Local-LDA \cite{brody2010unsupervised}, HFT \cite{mcauley2013hidden}, MF \cite{koren2009matrix,}, NVDM \cite{miao2016neural}, ProdLDA \cite{srivastava2017autoencoding}, and VRLDA.


\subsection{Interpretablity}

In Table \ref{tab:top-words}, we show the top 10 words in topics for the BeerAdvocate and the Yelp data. Words associated in sub-aspects are clearly related to each other. For example, in the beer dataset, the topic ``dark'' contains words such as ``black'', ``tans'', and ``brown''. Furthermore, we can see that the topics within every aspect are also correlated with one another. In the beer example, if we look at the topic neighbours of ``dark'', we can see the topic ``yellow''. Note that the ``dark'' and ``yellow'' topics are learned within the \emph{same} aspect in our model. The same pattern can be observed in the Yelp data where  we can recover topics corresponding to food types, such as ``Chinese'', ``Pizza'', and ``Breakfast''.


\subsection{Quantitative Assessment}

We perform a several quantitative evaluations of our model. We first demonstrate that we can successfully disentangle different aspects at the sentence level. We evaluate this on the two available annotated datasets: CitySearch \cite{ganu2009beyond} and BeerAdvocate \cite{mcauley2012learning}. Prior work on sentence aspect classification shows that Local-LDA is one of the most successful at capturing aspects in an unsupervised way \cite{lu2011multi}. Therefore, we compare against both LDA and Local-LDA. We also train fully supervised SVM
classifier\footnote{We use the SVM implementation in scikit-learn 0.19.2 \cite{scikit-learn}.} on the labeled data. As presented in Table \ref{tab:masl-results}, VALTA outperforms other approaches in terms of both accuracy and F1-score. 

Next, we quantitatively evaluate the top-words learned in topics. According to \textcite{lau2014machine}, NPMI is a good metric for qualitative evaluation of topics in terms of matching human judgment. We measure NPMI both at the sentence and the review level to take both aspects and topics into account. For every baseline, we only compared at the input-level that it was trained on. For example, LDA is trained at the review level while Local-LDA is trained at the sentence level. We report results in Table \ref{tab:NPMI-table}. We can see that VALTA performs significantly better than baselines, both at the sentence and review levels. Note that we keep the overall number topics for all other baselines to be the same as VALTA ($A \times K$). 


\subsection{Genre Discovery and Aspect-Based Analysis}

In Table \ref{tab:top-words}, we show that we can successfully learn a structured representation of aspects and topics. An interesting question is whether based on this representation, we have managed to cluster the items in a reasonable way. Furthermore, can we now perform an aspect-based comparison of different items? In this section, we investigate this question both qualitatively and quantitatively. We hypothesize that if the learned representation of the item is sufficiently rich in capturing the structure of the data, then even a simple classifier should be able to accurately distinguish between the categories. After training VALTA and other baselines, we fit a multi-class SVM to classify the items.\footnote{We use the SVC implementation in sklearn 0.19.2.} Results are shown in Figure \ref{fig:aspect-clustering} (left). VALTA outperforms other approaches with respect to clustering items in an unsupervised manner due to is structured nature. 

To inspect VALTA's ability to enable aspect-based comparison, we perform the following experiments: for both the BeerAdvocate and Yelp restaurant data, we manually select three categories of items that are different with respect to every aspect. We encode all the reviews associated with these items and we plot the histogram of parameters $\p_{i,u}$, as well as their kernel density estimation in different aspects. 

Figure \ref{fig:aspect-clustering} shows that item representations cluster appropriately within topics. For example, American Porter is a sweet dark beer, thus the histogram of $\p_{i,u}$ for American Porter is on the side of "dark" in the appearance aspect and on the side of "sweet" in the taste aspect. American IPA on the other hand is bitter pale beer, thus it has very little overlap with American Porter in either aspects. Note that in the beer example, we trained with $K = 2$. Since $\p_{i,u}$ are parameters of the Concrete distribution, we only need to see value for 1 as the other one provides no addition information.

\subsection{Recommendation Performance}

As noted in the methodology section, VALTA's generative model is also trained to predict rating for pairs of users and items, based on their aspect and topic representation. Results on MSE are shown in Table \ref{tab:recommender-results}. It can be observed that VALTA outperforms baselines in two of the datasets by taking both the rating and our structured review representation into account, and perform reasonably close to state-of-art (HFT) in other cases.




\begin{table}[!t]
    \centering
    \begin{tabular}{l|cccc|c}
        Dataset         & MF    & LDA    & HFT      & VRLDA  & VALTA  \\
        \hline
        \hline
        Beer            & 1.32  & 0.714       & 0.552 & 0.611 & \textbf{0.437}   \\
        Yelp            & 2.32  & 1.894       & \textbf{1.225}  & 1.540 & 1.236   \\ 
        Clothing        & 0.568 & 0.444       & \textbf{0.316} & 0.400 & \textbf{0.315}   \\
        Movies          & 0.242 & 0.217       & 0.197 & 0.201 & \textbf{0.154}   \\
        \hline
    \end{tabular}
    \caption{MSE comparison with baselines on the test data.}
    \label{tab:recommender-results}
\end{table}

\section{Conclusion}

We have proposed VALTA, a novel VAE-based model that instantiates structured probabilistic topic models in combination with an inference neural network to learn aspect-based representations of reviews. VALTA uncovers interpretable aspects, and additional structure (sub-aspects) beneath these. These representations enable one to measure similarity with respect to individual aspects, and thus perform aspect-wise clustering. Furthermore, we demonstrated the these representations afford improved generalization, as assessed in zero-shot settings. 

Our hope is that structured (disentangled) representations will see increased development and use in natural language processing (NLP) applications, as these may allow greater generalizability and transparency. 


%

\printbibliography

\newpage
\appendix
\onecolumn

\section{NPMI estimation}

We define the probability of word $i$;  $ p(w_{i})$, and the joint probability of two words $w_{i}$ and $w_{j}$ $p(w_{i}, w_{j})$ occurring together as their relative frequency. Let $N_{i}$ be the total number of documents where $w_{i}$ is present and $N_{i,j}$ be the total number documents where $w_{i}$ and $w_{j}$ are both present:  
\begin{align}
    &p(w_{i}) = \frac{\sum_{n=1}^{N}\mathds{1} \left[w_{i} \in x_{n} \right]}{\sum_{n=1}^{N}\mathds{1} \left[x_{n} \right]} =\frac{N_{i}}{N} \\
    &p(w_{i},w_{j}) = \frac{\sum_{n=1}^{N}\mathds{1} \left[w_{i},w_{j} \in x_{n} \right]}{\sum_{n=1}^{N}\mathds{1} \left[x_{n} \right]} =\frac{N_{i,j}}{N} 
\end{align}
NPMI is typically computed for top $T$ words for a given topic. The formula for computing NPMI for a given word $w_{i}$ is stated as:
\begin{equation}
    \textit{NPMI}(w_{i}) = \sum_{j}^{T-1}\frac{\log \frac{p(w_{i},w_{j})}{p(w_{i})p(w_{j})}}{-\log p(w_{i},w_{j})}
\end{equation}
In order to compute NPMI for a particular topic $k$, we compute this value for all $T$ top words associated to that topic:
\begin{equation}
    \textit{NPMI}_{k} = \frac{\sum_{i=1}^{T} \textit{NPMI}(w_{i} \in \textit{Topic}_{k})}{T}
\end{equation}
Finally, we compute the overall NPMI as the average of NPMIs for every topic:
\begin{equation}
    \textit{NPMI} = \frac{\sum_{k=1}^{K} \textit{NPMI}_{k}}{K}
\end{equation}


\section{Top Words}


In Figure \ref{app:tab:top-words}, we show the top 10 words for the datasets. 
\begin{table*}[!b]
    \centering
    \begin{tabular}{c|c|c||c|c||c|c||c}
    \multicolumn{3}{c||}{\textbf{Others}} & \multicolumn{2}{c||}{\textbf{General}} & \multicolumn{2}{c||}{\textbf{Genre}} & \multicolumn{1}{c}{\textbf{Awards}} \\
     \hline
     \hline
     dr             & colordeau     & dr        & movie & blu       & life      & war           & score\\
     frankenstein   & sorboone      & paris     & good  & sdh       & society   & parenthetical & perfroamnce \\
     grade          & jacques       & staci     & dvd   & audio     & humans    & khz           & performances\\
     self           & dr            & layne     & watch & ray       & world     & soldiers      & director \\
     wow            & universiy     & february  & film  & criterion & people    & poll          & role \\
     12             & dauphine      & self      & just  & ratio     & man       & troops        & directed \\
     mti            & versailles    & filmthe   & like  & grain     & jesus     & ship          & oscar \\
     enhancement    & ninja         & harp      & time  & dolby     & christ    & france        & cinemtrpgaphy \\
     06             & pantheon      & en        & really& dvd       & lives     & pacific       & filem \\
     stangleove     & turtles       & grade     & movies& transfer  & earth     & german        & quot \\
    \end{tabular}
    \caption{Top 10 words learned by VALTA: movies}
    \label{app:tab:top-words}
\end{table*}

\end{document}